\documentclass{nesy2026}

\usepackage{booktabs}
\usepackage{mathtools}
\usepackage{enumitem}
\usepackage{tikz}
\usetikzlibrary{arrows.meta,positioning,calc}
\definecolor{posterblue}{RGB}{8,24,96}
\definecolor{lightblue}{RGB}{235,243,250}
\definecolor{lightgreen}{RGB}{238,248,238}
\usepackage{xspace}
\usepackage{microtype}
\usepackage[capitalize,nameinlink]{cleveref}
\usepackage{algorithm}
\usepackage{algpseudocode}
\usepackage{xcolor}
\usepackage{listings}

\usepackage{tikz}
\usetikzlibrary{positioning,calc,arrows.meta}

\lstdefinelanguage{ProbLog}{
  morekeywords={
    nn,query,evidence,true,false
  },
  sensitive=true,
  morecomment=[l]{\%},
  morestring=[b]"
}

\lstdefinestyle{problogstyle}{
  language=ProbLog,
  basicstyle=\ttfamily\footnotesize,
  keywordstyle=\bfseries,
  commentstyle=\itshape\color{gray},
  columns=fullflexible,
  keepspaces=true,
  showstringspaces=false,
  breaklines=true,
  frame=single,
  rulecolor=\color{black!25},
  xleftmargin=1em,
  xrightmargin=1em,
  aboveskip=0.45em,
  belowskip=0.45em
}

\algrenewcommand\algorithmicrequire{\textbf{Require:}}
\algrenewcommand\algorithmicensure{\textbf{Ensure:}}

\newcommand{\DeepSWIP}{DeepSWIP\xspace}

\newcommand{\ProbLog}{ProbLog\xspace}
\newcommand{\DeepProbLog}{DeepProbLog\xspace}
\newcommand{\WMC}{\operatorname{WMC}}
\newcommand{\I}{\mathbb{I}}
\newcommand{\E}{\mathbb{E}}
\newcommand{\Prb}{\mathbb{P}}
\newcommand{\doop}{\operatorname{do}}
\newcommand{\mat}{\mathrm{mat}}

\title[DeepSWIP]{DeepSWIP: Quotient-WMC Counterfactuals for Neural Probabilistic Logic Programs}

\clearauthor{%
\Name{Saimun Habib}\Email{S.Habib-1@sms.ed.ac.uk}\\
\addr University of Edinburgh
\AND
Vaishak Belle\Email{vbelle@ed.ac.uk}\\
\addr University of Edinburgh
\AND
Fengxiang He\Email{fhe@ed.ac.uk}\\
\addr University of Edinburgh
}

\begin{document}
\maketitle

\begin{abstract}
Neurosymbolic systems such as \DeepProbLog combine neural perception with probabilistic logic, but standard inference is associational. Counterfactual reasoning additionally requires a causal semantics for interventions and evidence. We introduce \DeepSWIP, a single-world counterfactual semantics for \DeepProbLog programs. Using \emph{neural materialization}, we reduce fixed-context neural predicates to ordinary \ProbLog choices, apply Single World Intervention Programs (SWIPs), and compute counterfactuals by weighted model counting (WMC) over a single transformed program. Under finite grounding and unique-supported-model assumptions, \DeepSWIP is exact relative to the learned materialized FCM. The standard quotient-WMC form of \ProbLog conditionals identifies active neural probabilities and explains intervention cleaning, calibration sensitivity, and rare-evidence instability. Experiments on MPI3D confirm the transformation against a DeepTwin construction against 12,000 queries, as predicted and a 2.14$\times$ inference speedup from avoiding the Twin's endogenous duplication. A SUMO HOV experiment shows that neural calibration degradation biases plug-in estimates, while a correctly scoped randomized-policy AIPW estimator removes most first-order bias for population mean and ATE estimands. Code is at \url{https://github.com/saibib/deep_SWIP}
\end{abstract}

\section{Introduction}
\label{sec:intro}

In \DeepProbLog, Scallop, and related probabilistic logic programming (PLP) systems, neural modules map raw inputs to distributions over symbolic atoms, and symbolic inference composes these uncertain atoms through structured logical rules \citep{manhaeve2018deepproblog,li2023scallop}. This architecture is effective for prediction under perceptual uncertainty, but prediction is not intervention. A model that answers $\Prb(Y\mid X=x)$ cannot answer what would have happened under $\doop(X=x)$, nor what would have happened under $\doop(X=x)$ given contradictory factual evidence.

Counterfactual reasoning is central to explanation, robustness, fairness, and decision making \citep{pearl2009causality,halpern2016actual}. In PLPs, counterfactuals have been given causal semantics by translating programs into functional causal models (FCMs) and evaluating counterfactual queries through Twin Networks \citep{kiesel2023whatif}. The Twin construction is semantically convenient. It duplicates the endogenous structure into factual and counterfactual copies while sharing exogenous choices. However, duplication is expensive for relational programs and awkward for neural predicates \cite{habib2026swip}. In a visual \DeepProbLog program, should the convolutional predicate be evaluated twice, once per world, or shared as an exogenous symbolic uncertainty? More importantly, duplication obscures the single-world semantics of interventions that motivates Single World Intervention Graphs (SWIGs) and creates unnecessary (and unverifiable) independence assumptions \citep{richardson2013swigs,richardson2023singleworld, habib2026swip}.

The SWIP methodology of \citet{habib2026swip} translates SWIG-style intervention surgery into \ProbLog program rewriting. The methodology defines causal semantics for removing clauses defining intervened atoms, redirecting downstream uses to fixed intervention atoms, and querying the resulting single-world program to produce programs equivalent to the interventional and counterfactual distribution specified by the FCM. \DeepSWIP extends this idea to \DeepProbLog. Once neural predicates are materialized into probabilistic choices, counterfactual inference becomes a quotient of weighted model counts,
\begin{equation}
  Q(p)=\Prb(Y_x=y\mid E=e)=\frac{\Phi^x_{Y,E}(p)}{\Phi^x_E(p)},
  \label{eq:intro-quotient}
\end{equation}
where $p$ are materialized neural probabilities and $\Phi$ are multilinear WMC polynomials. This algebraic representation is the paper's organizing principle and explains why intervened mechanisms have zero derivative, why calibration error rather than classification accuracy is the relevant neural failure mode, and why rare evidence causes structural instability through the denominator $\Phi_E^x$.

Empirically, we evaluate DeepSWIP on two complementary settings. MPI3D tests the full neurosymbolic pipeline with neural predicates, symbolic rules, exact factor-level counterfactual ground truth, and a Twin-style comparison. The SUMO traffic experiments  tests the statistical side of the framework with imperfect neural traffic-state estimates, rare-evidence amplification, and the distinction between individual plug-in counterfactual prediction and population-level AIPW/DML estimation.

We make four contributions: (i) neural materialization for fixed-context \DeepProbLog counterfactuals; (ii) a correctness result lifting SWIP to the learned materialized FCM; (iii) a quotient-WMC analysis of active neural predicates, intervention cleaning, calibration sensitivity, and rare-evidence instability; and (iv) MPI3D and SUMO experiments separating exact symbolic inference from statistical error in learned predicates.

\section{Background and Related Work}
\label{sec:background}

\paragraph{Probabilistic logic and counterfactuals.}
A ground \ProbLog program consists of probabilistic facts and deterministic rules. Under distribution semantics, probabilistic facts define independent exogenous choices and the logic program maps each exogenous assignment to a unique symbolic world, provided the program has unique supported models \citep{deraedt2007problog}. \citet{kiesel2023whatif} show how to interpret such programs as FCMs and answer counterfactual queries by a Twin Network transformation, proving consistency with Pearl-style counterfactuals and CP-logic \citep{vennekens2009cplogic}. \citet{habib2026swip} instead proposes single-world intervention programs, following the SWIG idea that an intervention splits a node into a random factual part and a fixed intervention part \citep{richardson2013swigs}. Our work assumes these ordinary-\ProbLog results and asks how they extend when some exogenous symbolic choices are produced by neural predicates. 

\paragraph{Neural probabilistic logic and statistical correction.}
\DeepProbLog introduces neural predicates of the form $\operatorname{nn}(m,x,y)$, where a neural model $m$ maps input $x$ to a distribution over symbolic values $y$ \citep{manhaeve2018deepproblog}. Inference is typically reduced to arithmetic circuits or WMC after evaluating neural outputs. Scallop and related systems similarly reduce neural-symbolic inference to provenance circuits or weighted logical inference \citep{li2023scallop,valentin2025dpnl}. This circuit view is unified by algebraic model counting, which casts WMC, gradients, and related tasks as semiring evaluations over a shared circuit \citep{kimmig2017amc,maene2025gradient}; we use this perspective in \Cref{sec:theory}. \DeepSWIP is compatible with such solvers but our contribution does not focus on them. Instead, %we provide the causal transformation and quotient semantics of counterfactual queries after neural materialization. %In a separate thread of work, deep causal models learn generative mechanisms for counterfactual inference directly but lack the expressivity of first order logic \citep{pawlowski2020deep, xia2021causal, zecevic2021relating}. 
we treat neural predicates as probability estimators inside a causal logic program and this separation lets us analyze how neural nuisance error propagates through symbolic counterfactuals. To create causal estimands with desirable properties, we implement Double/debiased machine learning (DML) \citep{chernozhukov2018double,robins1994estimation}. We do not claim universal DML for arbitrary logic queries but we prove a plug-in sensitivity result for general \DeepSWIP queries and use AIPW only in experiments where the target is a randomized binary-treatment population estimand.

\section{Preliminaries}
\label{sec:preliminaries}

Let $\mathcal P=(\mathcal F,\mathcal R,\mathcal N)$ be a finite ground \DeepProbLog program, where $\mathcal F$ contains ordinary probabilistic choices, $\mathcal R$ deterministic rules, and $\mathcal N$ neural predicates. Let $\mathcal B$ be the finite Herbrand base induced by the query, evidence, intervention, and neural contexts under consideration. We write $\mathcal E\subseteq\mathcal B$ for the external atoms; ordinary probabilistic choices from $\mathcal F$ together with the categorical choices obtained by neural-materializing predicates in $\mathcal N$. The internal atoms are $\mathcal I=\mathcal B\setminus\mathcal E$ and are defined by the rules in $\mathcal R$. For an assignment \(u\in\{0,1\}^{\mathcal E}\) of external atoms, write
\[
  \operatorname{SM}_{\mathcal R}(u)
  =
  \{v\in\{0,1\}^{\mathcal I}: u\cup v \text{ is a supported model of }
  \mathcal R\}.
\] 
For an intervention \(X=x\), let \(\mathcal R_{x}\) denote the rule set obtained by deleting all rules whose head is in \(X\) and adding the corresponding constant assignments \(X:=x\). 

We assume for the query, evidence, intervention, and neural contexts under consideration, the grounding of \(\mathcal P\) is finite. Moreover, each neural predicate \(k\in\mathcal N\) has a finite output domain \(\mathcal Y_k=\{y_{k,1},\ldots,y_{k,r_k}\},\) and for fixed parameters \(\eta_k\) and context \(c_k\) returns a categorical distribution
\begin{equation}
    p_{k,j}(\eta_k,c_k) = \Prb_{\eta_k}(Y_k=y_{k,j}\mid c_k),
  \qquad
  \sum_{j=1}^{r_k}p_{k,j}(\eta_k,c_k)=1.
\label{eq:neural-probs}
\end{equation}
We write $p(\eta,c)$ for all ordinary and neural-materialized weights. We also require $\Prb_{M_{\eta,c}}(E=e)>0$ for every conditional query $\Prb(Y_x=y\mid E=e)$, so that the WMC quotient is defined.

The semantics of $\mathcal{P}$ define a probability distribution over the possible worlds of the assignments of probabilistic facts and neural-materialized choices as mutually independent external choices. Thus, for fixed \((\eta,c)\), every assignment \(u\in\{0,1\}^{\mathcal E}\) has probability
\[
  \Prb_{\eta,c}(u)
  =
  \prod_{a\in u} p_a(\eta,c)
  \prod_{a\in \mathcal E\setminus u}\bigl(1-p_a(\eta,c)\bigr),
\]
with annotated-disjunction choices interpreted as one-hot categorical choices. 

Finally, we assume unique supported models before and after intervention:
\[
  \left|\operatorname{SM}_{\mathcal R}(u)\right|=1,
  \qquad
  \left|\operatorname{SM}_{\mathcal R_x}(u)\right|=1
  \quad
  \text{for all } u\in\{0,1\}^{\mathcal E}.
\]
Equivalently, the materialized program induces a well-defined FCM. Acyclic rule dependency is sufficient but not necessary.

\section{The DeepSWIP Transformation}
\label{sec:method}
The purpose of DeepSWIP is to separate neural perception from symbolic counterfactual inference. A \DeepProbLog neural predicate is not intervened on as a neural network, rather, for fixed parameters and fixed input context, it induces a finite distribution over symbolic values. We therefore freeze the neural computation and expose only the induced symbolic uncertainty to the logic program. This turns a neural probabilistic logic program into an ordinary \ProbLog program, after which standard causal program transformations can be applied. 

\begin{definition}[Neural materialization]
\label{def:materialization}
Fix neural parameters $\eta$ and contexts $c$. The materialized program $\mathcal P_{\mat}(\eta,c)$ is the ordinary \ProbLog program obtained by replacing each neural predicate $\operatorname{nn}(m_k,c_k,Y_k)$ by the categorical probabilistic choice
\[
  p_{k,1}::a_k(y_{k,1});\cdots;p_{k,r_k}::a_k(y_{k,r_k}),
\]
where $p_{k,j}=p_{k,j}(\eta_k,c_k)$. The deterministic rule set $\mathcal R$ is unchanged.
\end{definition}

\begin{theorem}[Materialization equivalence]
\label{thm:materialization}
Fix $\eta$ and $c$. For every ground formula $\varphi$ over the symbolic atoms of $\mathcal P$,
\[
  \Prb_{\mathcal P}(\varphi\mid \eta,c)=\Prb_{\mathcal P_{\mat}(\eta,c)}(\varphi).
\]
\end{theorem}

Materialization gives the object on which intervention surgery is well-defined. The intervention itself is a program transformation that removes the mechanisms defining intervened atoms, redirects downstream rules to fixed intervention atoms, and leaves all non-intervened mechanisms unchanged. Thus, DeepSWIP inherits the single-world interpretation of SWIGs while retaining the ordinary distribution semantics of the materialized ProbLog program.
\begin{algorithm}[h]
\caption{DeepSWIP transformation}
\label{alg:deepswip}
\scriptsize
\KwIn{Ground DeepProbLog program $P=(F,R,N)$; neural parameters $\eta$; contexts $c$; intervention $do(X=x)$; evidence $E=e$; query $Y=y$}
\KwOut{Counterfactual probability $P(Y_x=y\mid E=e;\eta,c)$}
$P_{\mathrm{mat}}\gets F\cup R$\;
\ForEach{$nn(m_k,c_k,Y_k)\in N$}{
    Evaluate $m_k(c_k)$ and obtain $p_{k,j}=P_{\eta_k}(Y_k=y_{k,j}\mid c_k)$ for $j=1,\ldots,r_k$\;
    Replace $nn(m_k,c_k,Y_k)$ in $P_{\mathrm{mat}}$ by $p_{k,1}::a_k(y_{k,1});\cdots;p_{k,r_k}::a_k(y_{k,r_k})$\;
}
$S_x\gets$ probabilistic choices of $P_{\mathrm{mat}}$\;
\ForEach{clause $C=(h\leftarrow b_1,\ldots,b_\ell)$ in $R$}{
    \eIf{$h$ defines an intervened atom in $X$}{
        discard $C$\;
    }{
        replace each body occurrence of $X_i$ by $X_{i,\mathrm{fix}}(x_i)$\;
        add the rewritten clause to $S_x$\;
    }
}
\ForEach{intervention assignment $X_i=x_i$}{
    add the deterministic fixed assertion $X_{i,\mathrm{fix}}(x_i)$ to $S_x$\;
}
Let $\Delta_x$ be the Boolean theory induced by the transformed program\;
$\Phi^x_{Y,E}(p)\gets \mathrm{WMC}(\Delta_x\wedge Y=y\wedge E=e;p)$\;
$\Phi^x_E(p)\gets \mathrm{WMC}(\Delta_x\wedge E=e;p)$\;
\Return{$\Phi^x_{Y,E}(p)/\Phi^x_E(p)$}\;
\end{algorithm}
\begin{theorem}[DeepSWIP correctness against FCM]
\label{thm:correctness}
Assume $\mathcal P_{\mat}(\eta,c)$ has unique supported models, and let $M_{\eta,c}$ be its induced FCM. Let $\mathcal S_x(\mathcal P_{\mat})$ be the SWIP transformation under $\doop(X=x)$. For any query $Y=y$ and evidence $E=e$ with $\Prb_{M_{\eta,c}}(E=e)>0$,
\[
  \Prb_{\DeepSWIP}(Y_x=y\mid E=e;\eta,c)=\Prb_{M_{\eta,c}}(Y_x=y\mid E=e).
\]
\end{theorem}

The theorem is conditional on the learned/materialized FCM. It says that symbolic counterfactual inference is exact given the materialized neural probabilities. It does not say that $\eta$ is the true data-generating parameter. Algorithm \ref{alg:deepswip} details the implementation of the program transformation and counterfactual querying.

We give a minimal example that illustrates the complete DeepSWIP pipeline. Consider a DeepProbLog traffic-monitoring fragment in which a neural predicate classifies an upstream sensor image into a latent traffic state, and symbolic rules derive congestion and delay:
\begin{lstlisting}[style=problogstyle]
nn(state_net, img42, [free,queued])::traffic_state(S).

0.20::rain.
0.10::roadworks.

incident  :- rain, roadworks.
congested :- traffic_state(queued).
congested :- incident.
delayed   :- congested.
reroute   :- incident.
\end{lstlisting}

For fixed neural parameters and input image \texttt{img42}, neural materialization evaluates the network once and replaces the neural predicate with its softmax output as an annotated disjunction. For example, if the network returns probability $0.30$ for \texttt{free} and $0.70$ for \texttt{queued}, the materialized program is now an ordinary ProbLog program containing \texttt{0.30::traffic\_state(free); 0.70::traffic\_state(queued).}
% \begin{lstlisting}[style=problogstyle]
% 0.30::traffic_state(free); 0.70::traffic_state(queued).
% \end{lstlisting}
% so the full materialized ProbLog fragment is
% \begin{lstlisting}[style=problogstyle]
% 0.30::traffic_state(free); 0.70::traffic_state(queued).

% 0.20::rain.
% 0.10::roadworks.

% incident  :- rain, roadworks.
% congested :- traffic_state(queued).
% congested :- incident.
% delayed   :- congested.
% reroute   :- incident.
% \end{lstlisting}

Now, suppose the factual evidence is that congestion was observed while it was raining, but the counterfactual query asks what the delay would have been under $do(\texttt{congested}=\texttt{false})$. 
% \[
%   \Prb\{\texttt{delayed}_{\texttt{congested=false}}
%   \mid \texttt{congested, rain}\}.
% \]
The counterfactual program is now:
\begin{lstlisting}[style=problogstyle]
0.30::traffic_state(free); 0.70::traffic_state(queued).

0.20::rain.
0.10::roadworks.

incident :- rain, roadworks.

congested_obs :- traffic_state(queued).
congested_obs :- incident.

0.0::congested_fix_true.
delayed :- congested_fix_true.

reroute :- incident.

evidence(congested_obs, true).
evidence(rain, true).
query(delayed).
\end{lstlisting}
The causal pathways between $\texttt{traffic\_state(queued)}\rightarrow\texttt{congested\_obs}$ and $\texttt{incident}\rightarrow\texttt{congested\_obs}$ remain in the denominator of the WMC quotient. They perform the abduction step by reweighting the exogenous explanations of the factual congestion, including the neural-materialized traffic-state choice. Only the outgoing use of \texttt{congested} in the descendant rule for \texttt{delayed} is redirected to the fixed false intervention.

The counterfactual probability is then computed as the corresponding WMC quotient with numerator additionally conjoining \texttt{delayed} and denominator conditioning on \texttt{congested\_obs} and \texttt{rain}. As desired, the query evaluates to zero because \texttt{delayed} cannot be derived through the intervened congestion mechanism, even though factual evidence says congestion occurred.

By contrast, \(\Prb_{\mathrm{DeepSWIP}}(\texttt{reroute}\mid \texttt{congested},\doop(\texttt{congested}=\texttt{false}))\) is still governed by \texttt{reroute :- incident}. Thus \DeepSWIP does not erase factual causes of the intervened variable, instead it only severs the intervened atom's defining equations and redirects downstream uses to the fixed value. In general, we can bound how much the materialized program changes.

% The counterfactual probability is then computed as the WMC quotient:
% % \[
% % \frac{
% % \operatorname{WMC}\!\left(
% % \Delta_{\mathrm{SWIP}}
% % \wedge
% % \texttt{delayed}
% % \wedge
% % \texttt{congested\_obs}
% % \wedge
% % \texttt{rain}
% % \right)}
% % {
% % \operatorname{WMC}\!\left(
% % \Delta_{\mathrm{SWIP}}
% % \wedge
% % \texttt{congested\_obs} 
% % \wedge
% % \texttt{rain}
% % \right)}.
% % \]

% \begin{equation}
%     \frac{
% \operatorname{WMC}\!\left(
% \Delta_{\mathrm{SWIP}}
% \wedge
% \texttt{delayed}
% \wedge
% \texttt{congested\_obs}
% \wedge
% \texttt{rain}
% \right)}
% {
% \operatorname{WMC}\!\left(
% \Delta_{\mathrm{SWIP}}
% \wedge
% \texttt{congested\_obs} 
% \wedge
% \texttt{rain}
% \right)}.
% \end{equation}
% As desired, this query evaluates to zero as \texttt{delayed} cannot be derived through the intervened congestion mechanism even though factual evidence says that congestion occurred.

% By contrast, $\Prb_{\mathrm{DeepSWIP}}(\texttt{reroute}
% \mid \texttt{congested},\mathrm{do}(\texttt{congested}=\texttt{false})).$ is still governed by the unchanged rule \texttt{reroute :- incident} and the unchanged upstream mechanism \texttt{incident :- rain, roadworks}. Therefore, DeepSWIP does not erase the factual causes of the intervened variable and only severs the defining equations of the intervened atom and redirects its downstream uses to the fixed intervention value. If no evidence or query refers to the factual copy of the intervened atom, this observed component is irrelevant and can be pruned. In general, we can bound how much the materialized program changes. 
\begin{proposition}[Program growth]
\label{prop:growth}
Let $\mathcal P_{\mat}$ have $m$ probabilistic choices and $|\mathcal R|$ deterministic clauses. For an intervention on $k$ ground atoms, $\mathcal S_x(\mathcal P_{\mat})$ contains at most $m+k$ probabilistic choices if fixed atoms are encoded as degenerate facts, and $m$ otherwise. Its number of clauses is at most $|\mathcal R|+k$ before pruning. 
\end{proposition}

\label{rem:growth-scope}
Proposition \ref{prop:growth} is a syntactic statement. WMC can still be exponential in circuit size or treewidth. 

\section{Quotient-WMC Semantics and Consequences}
\label{sec:theory}

The quotient \eqref{eq:quotient} is an instance of \emph{algebraic model counting} (AMC), which generalizes weighted model counting from the probability semiring to arbitrary commutative semirings, unifying a range of inference tasks as one circuit evaluation \citep{kimmig2017amc}. Both $\Phi^x_{Y,E}$ and $\Phi^x_E$ are AMC evaluations over the probability semiring on the single SWIP-transformed circuit $\Delta^x$, so the counterfactual conditional is a ratio of two AMC values on the same circuit. This view is more than terminological. The local sensitivities of Theorems \ref{thm:sensitivity} and \ref{thm:rare} are exactly the quantities computed by AMC over the \emph{gradient semiring}, the construction used to train neural predicates in DeepProbLog, and recently shown to unify gradient and model-counting computations in a single algebraic framework \citep{maene2025gradient}.

Consequently $\nabla Q$ is not only characterized analytically by \eqref{eq:quotient-derivative} but is computable by one gradient-semiring pass over $\Delta^x$. Intervention cleaning (\Cref{thm:cleaning}) is likewise algebraic with SWIP removing the intervened literals from the labeling, so their semiring weights never enter the count, and the active neural set is the support of the labeling on the pruned circuit. We read the results below as a quotient-AMC account of neurosymbolic counterfactuals and the causal SWIP transformation and the counterfactual reading of the quotient are our additions on top of the AMC machinery.

Let $\Delta_x$ be the Boolean theory induced by the SWIP-transformed materialized program. For a world $\omega$ over probabilistic choices $p_1,\ldots,p_m$, define
\begin{equation}
  w_p(\omega)=\prod_{i:\omega_i=1}p_i\prod_{i:\omega_i=0}(1-p_i),\qquad
  \WMC(\Gamma;p)=\sum_{\omega\models\Gamma}w_p(\omega).
\end{equation}
The previous section establishes that \DeepSWIP produces the correct transformed program. After materialization and intervention surgery, neural dependence enters only through the weights $p$, while the logic determines which assignments satisfy query and evidence. Thus a counterfactual conditional is a quotient of two weighted model counts. This quotient is the central algebraic object of the paper.

\begin{theorem}[Quotient-polynomial WMC semantics]
\label{thm:quotient}
For any query $Y=y$ and evidence $E=e$ with $\WMC(\Delta_x\wedge E=e;p)>0$,
\begin{equation}
  Q(p)=\Prb_{\Delta_x}(Y=y\mid E=e)
  =\frac{\Phi^x_{Y,E}(p)}{\Phi^x_E(p)},
  \label{eq:quotient}
\end{equation}
where
\[
  \Phi^x_{Y,E}(p)=\WMC(\Delta_x\wedge Y=y\wedge E=e;p),\qquad
  \Phi^x_E(p)=\WMC(\Delta_x\wedge E=e;p).
\]
Both $\Phi^x_{Y,E}$ and $\Phi^x_E$ are multilinear polynomials in the materialized probabilities.
\end{theorem}
The remaining results are consequences of this quotient representation. The transformation changes which weights appear in the numerator and denominator and this gives a clean lens to view intervention, pruning, calibration, and rare event stability through. 

\begin{theorem}[Intervention cleaning]
\label{thm:cleaning}
If a materialized probability $p_k$ appears only in clauses defining an intervened atom removed by SWIP, then $Q$ does not depend on $p_k$. Hence $\partial Q/\partial p_k=0$ wherever the derivative is defined.
\end{theorem}
This formalizes the intuition that intervened mechanisms are no longer active sources of uncertainty for the counterfactual query. Neural errors in a mechanism that has been surgically replaced cannot affect the resulting quotient. In effect, we should only consider the active neural set. 
\begin{proposition}[Active neural set]
Let \(Q(p)=\Phi^x_{Y,E}(p)/\Phi^x_E(p)\) be the DeepSWIP quotient.
If a materialized neural probability \(p_k\) does not occur in either
\(\Phi^x_{Y,E}\) or \(\Phi^x_E\) after SWIP and relevance pruning, then
\[
    \partial Q/\partial p_k=0.
\]
\end{proposition}
Thus each query induces an active set of neural probabilities and only probabilities that survive intervention surgery and appear in the numerator or denominator can influence the answer. It follows that the query is sensitive to the materialized neural probabilities and the logical rules and we can decompose this sensitivity accordingly
\begin{theorem}[Local calibration sensitivity]
\label{thm:sensitivity}
Let $p^\star$ be ideal materialized probabilities and assume $\Phi^x_E(p^\star)>0$. For $\hat p=p^\star+\varepsilon$ sufficiently close to $p^\star$,
\begin{equation}
  Q(\hat p)-Q(p^\star)=\nabla Q(p^\star)^\top\varepsilon+O(\|\varepsilon\|^2),
  \label{eq:taylor}
\end{equation}
where for each active coordinate
\begin{equation}
  \frac{\partial Q}{\partial p_k}(p)=
  \frac{1}{\Phi_E^x(p)}
  \left[
  \frac{\partial \Phi^x_{Y,E}}{\partial p_k}(p)-Q(p)\frac{\partial \Phi^x_E}{\partial p_k}(p)
  \right].
  \label{eq:quotient-derivative}
\end{equation}
\end{theorem}
The leading error term is a product of neural probability error and logical sensitivity. This is why calibration, rather than classification accuracy alone, is the relevant neural property for counterfactual reliability. A classifier may have high accuracy but poor probability calibration and such calibration error can still be amplified by the symbolic quotient. This also reveals a structural instability: when the evidence probability is small, small
neural calibration errors may induce large counterfactual errors.

\begin{theorem}[Rare-evidence instability]
\label{thm:rare}
Let $\delta=\Phi^x_E(p)>0$. If $|\partial \Phi^x_{Y,E}/\partial p_k|\le C_1$ and $|\partial \Phi^x_E/\partial p_k|\le C_2$ in a neighborhood of $p$, then
\[
  \left|\frac{\partial Q}{\partial p_k}(p)\right|\le \frac{C_1+C_2}{\delta}.
\]
Thus local sensitivity can grow as $1/\Prb(E=e)$ and diverge.
\end{theorem}
Rare-evidence instability is a property of the counterfactual functional itself. It is not caused by a particular neural architecture or optimizer. Better neural training can reduce $\|\hat p-p^\star\|$, but it cannot remove the denominator $\Phi_E^x(p)$ from the counterfactual query. This motivates the use of double machine learning in the context of neurosymbolic counterfactuals.

\begin{proposition}[Cross-fitting and scoped DML]
\label{prop:dml}
Let $A$ be the active set after intervention cleaning and pruning. For any plug-in estimator $\bar p$ of $p^\star$,
\[
  \E[Q(\bar p)-Q(p^\star)]
  =\nabla_A Q(p^\star)^\top \E[\bar p_A-p_A^\star]+O(\E\|\bar p_A-p_A^\star\|^2).
\]
\end{proposition}

Therefore calibration or cross-fitting can reduce first-order plug-in bias only to the extent that it reduces active probability bias. This is not a universal DML guarantee. AIPW/DML requires a target estimand, valid treatment/propensity structure, and a Neyman-orthogonal score. The proposition separates two claims that should not be conflated.

Full proofs are provided in \appendixref{app:proofs}. The main point for experiments is that \DeepSWIP has two separable error sources: exact symbolic inference relative to $p$, and statistical/perceptual error in estimating $p$.

\section{Experiments}
\label{sec:experiments}
\begin{table}[b]
\centering
\caption{
\textbf{MPI3D visual counterfactuals.}
}
\label{tab:mpi3d-visual}
\small
\begin{tabular}{lrrrr}
\toprule
\textbf{Component} & \textbf{Metric} & \textbf{Value} & \textbf{Comparison} & \textbf{Value} \\
\midrule
Shape predicate & Accuracy & 0.889 & Calibration gap & --0.046 \\
Size predicate  & Accuracy & 0.987 & Calibration gap & --0.026 \\
Colour predicate & Accuracy & 1.000 & Calibration gap & --0.001 \\
\midrule
DeepSWIP vs DeepTwin & Agreement & 100.0\% & Queries & 12{,}000 \\
Inference time & DeepSWIP & 3.57 ms & DeepTwin & 7.65 ms \\
Runtime ratio & Twin/SWIP & 2.14$\times$ &  &  \\
\midrule
\texttt{can\_roll} & RMSE & 0.000 & Bias & 0.000 \\
\texttt{can\_stack} & RMSE & 0.058 & Bias & 0.001 \\
\texttt{risky\_on\_shelf} & RMSE & 0.082 & Bias & --0.003 \\
\texttt{stable} & RMSE & 0.000 & Bias & 0.000 \\
\bottomrule
\end{tabular}
\end{table}
Our experiments are designed around the theory rather than breadth of applications. MPI3D tests visual neural materialization, the speedup from the single-world transformation, pruning, and intervention cleaning, using DeepSWIP--DeepTwin agreement as an implementation correctness check and factor labels as counterfactual ground truth. SUMO HOV tests calibration-sensitive plug-in error and scoped AIPW correction.

\subsection{MPI3D visual counterfactuals}
\label{sec:mpi3d}
\paragraph{Setup.}
MPI3D-toy is factorized by color, shape, size, camera, background, and pose \citep{gondal2019transfer}. We train neural predicates for shape, size, and color, materialize  and use symbolic rules for \texttt{can\_roll}, \texttt{can\_stack}, \texttt{stable}, and \texttt{risky\_on\_shelf}. Counterfactuals intervene on shape or size and compare \DeepSWIP with a DeepTwin construction over the same materialized program. Factor labels supply ground-truth potential outcomes.

% \begin{table}[t]
% \centering
% \caption{
% \textbf{Scoped DML/AIPW sanity check on MPI3D.}
% This is a separate randomized size-intervention experiment with a valid population ATE estimand. The visual shape-intervention counterfactuals in Table~\ref{tab:mpi3d-visual} are not DML estimators. Bias is measured relative to the exact MPI3D factor-ground-truth ATE.
% }
% \label{tab:mpi3d-dml}
% \small
% \begin{tabular}{llrr}
% \toprule
% \textbf{Query} & \textbf{Estimator} & \textbf{ATE estimate} & \textbf{Bias} \\
% \midrule
% \texttt{can\_stack}
% & True ATE & --0.3333 & 0.0000 \\
% & Naive plug-in & --0.3242 & +0.0091 \\
% & Cross-fitted plug-in & --0.3525 & --0.0191 \\
% & AIPW/DML & --0.3368 & --0.0034 \\
% & Difference-in-means & --0.3361 & --0.0027 \\
% \midrule
% \texttt{risky\_on\_shelf}
% & True ATE & +0.6667 & 0.0000 \\
% & Naive plug-in & +0.6277 & --0.0390 \\
% & Cross-fitted plug-in & +0.6825 & +0.0159 \\
% & AIPW/DML & +0.6710 & +0.0043 \\
% & Difference-in-means & +0.6717 & +0.0050 \\
% \bottomrule
% \end{tabular}
% \end{table}

\paragraph{Results.}
With 8,368 training and 2,000 test images, neural predicate accuracies were 0.889 for shape, 0.987 for size, and 1.000 for color. Table~\ref{tab:mpi3d-visual} reports across 500 evaluation images, six shape interventions, and four queries (12,000 ProbLog calls), \DeepSWIP and DeepTwin agreed on 100\% of queries. \DeepSWIP averaged 3.57 ms/query and DeepTwin 7.65 ms/query, a $2.14\times$ speedup. Ground-truth RMSE was zero for intervention-determined queries and small for queries depending on non-intervened predicates: 0.058 for \texttt{can\_stack} and 0.082 for \texttt{risky\_on\_shelf}. Pruning under $\doop(\texttt{shape=sphere})$ reduced statement counts from 18 to 5, 8, 11, and 7 for the four queries, pruning color in all cases.

DeepSWIP and DeepTwin are both exact relative to the materialized FCM (\Cref{thm:correctness}), so we use their agreement as a correctness check on the transformation rather than as evidence for the semantics. Across 500 evaluation images, six shape interventions, and four queries (12{,}000 ProbLog calls) they agreed on 100\% of queries, exactly as the equivalence theorem predicts, and any disagreement would have indicated a transformation or implementation bug. The substantive results are the speedup and the propagation of neural error. Avoiding the Twin's endogenous duplication (\Cref{prop:growth}) reduced inference from 7.65 to 3.57\,ms/query, a 2.14$\times$ speedup on this program, consistent with the duplication argument. Because DeepSWIP is exact, the only error against factor ground truth comes from neural calibration on the \emph{active} predicates: RMSE was zero for intervention-determined queries (\texttt{can\_roll}, \texttt{stable}), empirically exhibiting intervention cleaning, and small for queries depending on non-intervened predicates (0.058 for \texttt{can\_stack}, 0.082 for \texttt{risky\_on\_shelf}). Pruning under \texttt{do(shape=sphere)} reduced statement counts from 18 to 5, 8, 11, and 7, removing color in all cases. Neural predicate accuracies were 0.889 (shape), 0.987 (size), and 1.000 (color).

\subsection{HOV randomized-policy calibration and DML stress test}
\label{sec:hov}
\paragraph{Setup.}
The HOV experiment uses paired SUMO simulations of restricted and open-lane policies, yielding potential outcomes $Y_i(0)$ and $Y_i(1)$ for throughput \citep{lopez2018microscopic}. We construct a randomized-policy observation process with $T_i\sim\operatorname{Bernoulli}(0.5)$ and target $\psi_1=\E[Y(1)]$ and $\tau=\E[Y(1)-Y(0)]$. A neural traffic-state predicate is trained under increasing sensor noise. The plug-in \DeepSWIP nuisance model maps state probabilities to state-conditional throughput means, while AIPW uses known propensity $g=0.5$ and fold-specific observed-only nuisance estimates. AIPW is evaluated as a population estimator, not an individual counterfactual RMSE.

The HOV experiment uses paired SUMO simulations of restricted and open-lane policies, yielding potential outcomes $Y_i(0)$ and $Y_i(1)$ for throughput \citep{lopez2018microscopic}. To test valid AIPW/DML, we construct a randomized-policy observation process: $T_i\sim\operatorname{Bernoulli}(0.5)$ and $Y_i=T_iY_i(1)+(1-T_i)Y_i(0)$. The target estimands are $\psi_1=\E[Y(1)]$ and $\tau=\E[Y(1)-Y(0)]$. A neural traffic-state predicate is trained under increasing sensor noise. The plug-in \DeepSWIP nuisance model maps state probabilities to state-conditional throughput means.  AIPW is evaluated as a population estimator, not an individual counterfactual RMSE.
\begin{figure}[h]
    \centering
    \vspace{-0.75em}
    \includegraphics[width=0.85\linewidth]{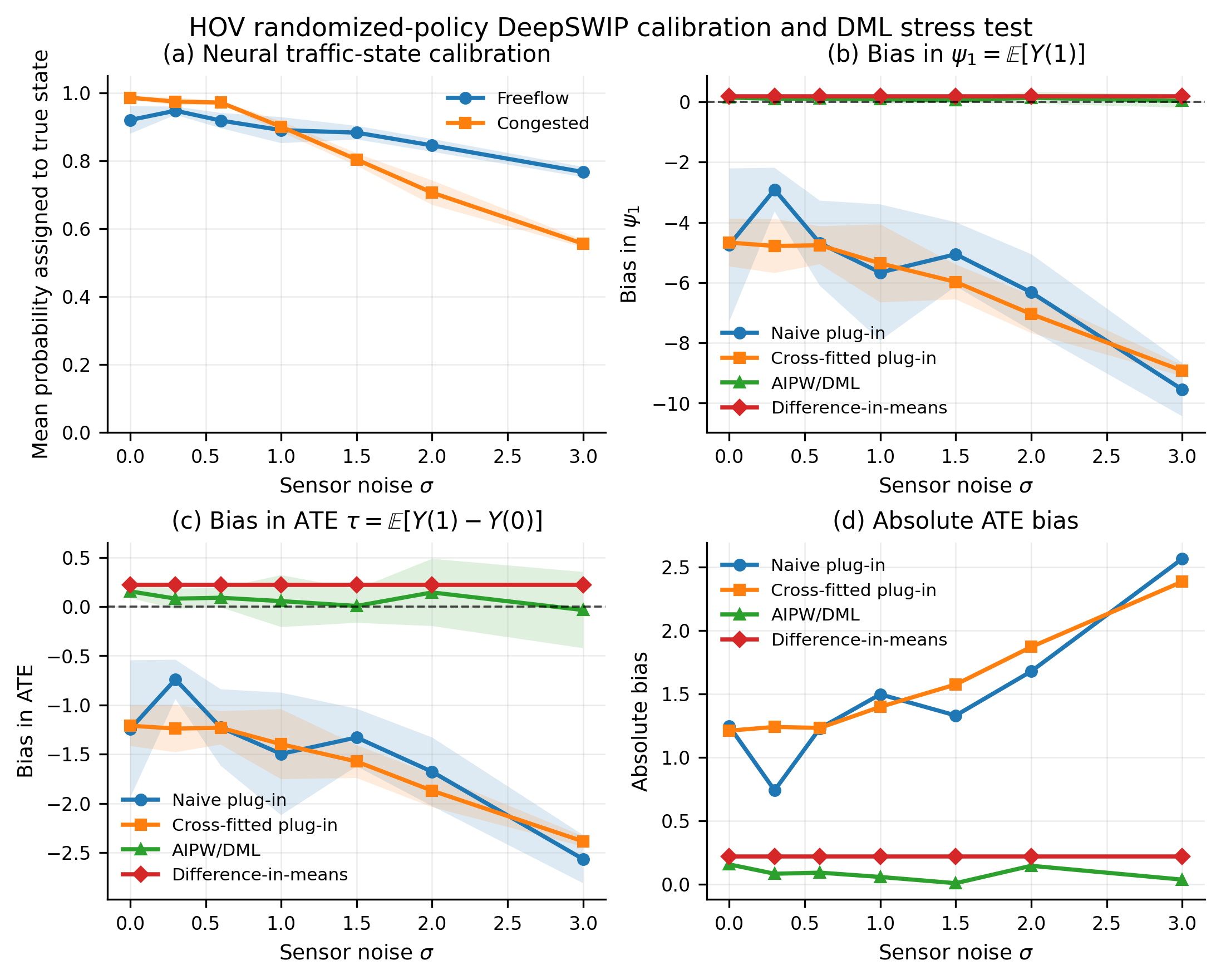}
    \vspace{-0.75em}
    \caption{
    \textbf{HOV randomized-policy calibration and DML stress test.}
    }
    \label{fig:hov-dml-paper}
    \vspace{-1.0em}
\end{figure}

\paragraph{Results.}
The dataset has $N=5000$ episodes: 4247 freeflow and 753 congested. The true values are $\E[Y(1)]=248.221$, $\E[Y(0)]=184.893$, and $\tau=63.329$. Figure \ref{fig:hov-dml-paper} shows, as sensor noise increases, congested calibration falls from 0.985 to 0.556. Naive plug-in bias worsens for both $\psi_1$ ($-4.75$ to $-9.55$) and ATE ($-1.24$ to $-2.57$), while cross-fitted plug-in remains biased. AIPW/DML stays near zero bias across noise levels, as does difference-in-means under randomized treatment.

\section{Discussion and Limitations}
\label{sec:discussion}

\DeepSWIP's exactness is exactness relative to the learned materialized FCM. If a neural predicate is miscalibrated, the symbolic counterfactual can be exactly wrong with respect to the external data-generating process. This is not a defect of SWIP; it is the distinction between causal-symbolic inference and statistical estimation. The quotient derivative in \Cref{thm:sensitivity} makes this distinction apparent as only active, non-intervened neural probabilities matter, and their errors are amplified by the logic and by $1/\Phi_E$.

The DML result is deliberately scoped. The HOV AIPW experiment is valid because we construct a randomized binary treatment with known propensity and evaluate population means/ATEs. This does not imply that arbitrary \DeepSWIP counterfactual queries have an orthogonal score, nor that AIPW estimates individual counterfactual outcomes. For general path-specific, continuous, or relational counterfactuals, deriving the efficient influence function is a separate statistical problem. Similarly, approximate WMC methods may accelerate inference, but approximation error in a quotient requires denominator-aware control; we leave approximate counterfactual WMC to future work.

Finally, our MPI3D rules are intentionally simple. The experiment is not a claim about state-of-the-art vision or physics. Its role is to breackdown the NeSy counterfactuals as  neural predicates are materialized, symbolic interventions are performed in a single world, and WMC computes the resulting quotient exactly. 

\section{Conclusion}
\label{sec:conclusion}

We presented \DeepSWIP, a framework for counterfactual reasoning in neural probabilistic logic programs. The method evaluates neural predicates once, materializes them as ordinary probabilistic choices, applies a single-world intervention transformation, and computes the result by WMC. This yields exact counterfactuals relative to the learned materialized FCM and exposes the counterfactual query as a quotient of WMC polynomials. The quotient view provides a common explanation for program growth, pruning, intervention cleaning, calibration sensitivity, rare-evidence instability, and the limited role of DML. Experiments on MPI3D and SUMO support these claims while keeping the statistical scope explicit. \DeepSWIP is therefore best understood not as a universal causal discovery method, but as a precise counterfactual inference layer for NeSy programs with explicit causal structure and learned probabilistic mechanisms.

\acks{Funding for this research was provided by NERC through an E4 DTP studentship (NE/S007407/1).}

\bibliography{deepswip_nesy2026}
\appendix

\section{Proofs}
\label{app:proofs}

\subsection{Proof of \Cref{thm:materialization}}
Fix $\eta$ and $c$. Each neural predicate $k$ induces a categorical distribution over a finite symbolic choice $U_k\in\mathcal Y_k$ with probabilities $p_{k,j}(\eta_k,c_k)$. Materialization replaces this random choice by an ordinary probabilistic choice with exactly the same probabilities. Ordinary probabilistic facts are unchanged. Therefore every complete assignment $u$ to all ordinary and neural-materialized choices has the same probability before and after materialization:
\[
  \Prb_{\mathcal P}(U=u\mid \eta,c)=\Prb_{\mathcal P_{\mat}(\eta,c)}(U=u).
\]
For a fixed $u$, the deterministic rules are identical in $\mathcal P$ and $\mathcal P_{\mat}$. By unique supported models, the closure of the rules under $u$ is unique and the same in both programs. Let $\Omega_\varphi$ be the set of assignments whose closure satisfies $\varphi$. Then
\[
\Prb_{\mathcal P}(\varphi\mid\eta,c)=\sum_{u\in\Omega_\varphi}\Prb_{\mathcal P}(u\mid\eta,c)=\sum_{u\in\Omega_\varphi}\Prb_{\mathcal P_{\mat}}(u)=\Prb_{\mathcal P_{\mat}}(\varphi).
\]

\subsection{Proof of \Cref{thm:correctness}}
By \Cref{thm:materialization}, the fixed-parameter \DeepProbLog program and $\mathcal P_{\mat}$ induce the same distribution over symbolic worlds. Since $\mathcal P_{\mat}$ has unique supported models, it induces an FCM $M_{\eta,c}$ whose exogenous variables are the probabilistic choices and whose endogenous structural assignments are the deterministic closures of the rules. The SWIP transformation removes clauses defining $X$, redirects downstream body literals to fixed atoms, and asserts the intervention values. This is exactly the program-level implementation of replacing the structural assignments for $X$ by constants while leaving other mechanisms unchanged. The ordinary-\ProbLog SWIP correctness theorem of \citet{habib2026swip}, applied to $\mathcal P_{\mat}$, gives equality between the probability computed on $\mathcal S_x(\mathcal P_{\mat})$ and the FCM counterfactual in $M_{\eta,c}$. The condition $\Prb(E=e)>0$ ensures the conditional probability is defined.

\subsection{Proof of \Cref{prop:growth}}
Materialization yields $m$ probabilistic choices. SWIP does not duplicate any retained rule. For each intervened atom, all clauses defining that atom are removed; all remaining clauses are either unchanged or have body literals renamed to fixed atoms. This operation cannot increase the number of retained original clauses. SWIP then adds at most one fixed assertion per intervened atom. If these assertions are deterministic facts they add no probabilistic choices; if encoded as degenerate probabilistic facts they add at most $k$. Hence the bounds $m+k$ and $|\mathcal R|+k$. A Twin construction creates factual and counterfactual endogenous copies, giving two copies of each endogenous rule before intervention-specific deletions. Exogenous choices may be shared, but endogenous symbolic structure is duplicated.

\subsection{Proof of \Cref{thm:quotient}}
Conditional probability gives
\[
\Prb_{\Delta_x}(Y=y\mid E=e)=\frac{\Prb_{\Delta_x}(Y=y,E=e)}{\Prb_{\Delta_x}(E=e)}.
\]
Under WMC, these probabilities are exactly $\WMC(\Delta_x\wedge Y=y\wedge E=e;p)$ and $\WMC(\Delta_x\wedge E=e;p)$. For any formula $\Gamma$,
\[
\WMC(\Gamma;p)=\sum_{\omega\models\Gamma}\prod_{i:\omega_i=1}p_i\prod_{i:\omega_i=0}(1-p_i).
\]
Each summand contains each $p_i$ at most once, either as $p_i$ or $1-p_i$. Expanding yields a multilinear polynomial. Thus both numerator and denominator are multilinear polynomials and the query is their quotient.

\subsection{Proof of \Cref{thm:cleaning}}
If $p_k$ appears only in clauses defining an intervened atom, those clauses are removed by SWIP. Therefore the transformed theory $\Delta_x$ contains no dependence of $Y$ or $E$ on the Boolean choice weighted by $p_k$. In the WMC sum, assignments to that variable factor as $p_k+(1-p_k)=1$ in both numerator and denominator. Thus neither polynomial depends on $p_k$, and the quotient derivative is zero.

\subsection{Proof of \Cref{thm:sensitivity}}
By \Cref{thm:quotient}, $Q(p)=N(p)/D(p)$ with $N=\Phi^x_{Y,E}$ and $D=\Phi^x_E$. Since $N$ and $D$ are polynomials and $D(p^\star)>0$, $Q$ is differentiable in a neighborhood of $p^\star$. Taylor's theorem gives \eqref{eq:taylor}. The quotient rule gives
\[
\frac{\partial Q}{\partial p_k}=\frac{D\,\partial N/\partial p_k-N\,\partial D/\partial p_k}{D^2}
=\frac1{D}\left[\frac{\partial N}{\partial p_k}-Q\frac{\partial D}{\partial p_k}\right],
\]
which is \eqref{eq:quotient-derivative}.

\subsection{Proof of \Cref{thm:rare}}
Using \eqref{eq:quotient-derivative} and $0\le Q(p)\le1$,
\[
\left|\frac{\partial Q}{\partial p_k}\right|
\le \frac1{\Phi_E^x(p)}\left(\left|\frac{\partial\Phi^x_{Y,E}}{\partial p_k}\right|+|Q(p)|\left|\frac{\partial\Phi_E^x}{\partial p_k}\right|\right)
\le \frac{C_1+C_2}{\delta}.
\]
Thus the derivative bound diverges as $\delta\to0$.

\subsection{Proof of \Cref{prop:dml}}
Partition worlds as $(\omega_A,\omega_{\bar A})$, where $A=A_x(Y,E)$. By definition, variables outside $A$ have no directed path to $Y$ or $E$ in the transformed dependency graph. Under the stated constraint-separation assumption, changing $\omega_{\bar A}$ cannot affect satisfaction of $Y$, $E$, or the retained clauses. Hence
\[
\Phi^x_{Y,E}(p)=\sum_{\omega_A}\I\{\omega_A\models\Delta_x^A\wedge Y\wedge E\}w_{p_A}(\omega_A)\sum_{\omega_{\bar A}}w_{p_{\bar A}}(\omega_{\bar A}).
\]
The inner sum is one because it sums the probability mass of independent removed choices. Therefore $\Phi^x_{Y,E}(p)=\Phi^{x,A}_{Y,E}(p_A)$. The same argument gives $\Phi^x_E(p)=\Phi^{x,A}_E(p_A)$, and taking quotients proves the claim.
Now, apply \Cref{thm:sensitivity} to $\bar p=p^\star+\varepsilon$. Coordinates outside $A$ are absent or have zero derivative by \Cref{thm:cleaning} and the above. Hence
\[
Q(\bar p)-Q(p^\star)=\nabla_AQ(p^\star)^\top(\bar p_A-p_A^\star)+O(\|\bar p_A-p_A^\star\|^2).
\]
Taking expectations yields the stated expansion. The final statement follows from the definition of DML: cross-fitting controls nuisance overfitting, but first-order bias is removed only when the estimator is built from a Neyman-orthogonal score for a valid target estimand.

\section{Experimental DeepProbLog Programs}
\label{app:deepswip-programs}

This appendix gives the DeepProbLog-style program templates used in the experiments.  The actual experiments first evaluate the neural predicates at a fixed input context and then replace them by ordinary probabilistic choices before applying the DeepSWIP transformation.  Thus the code below specifies the neural-symbolic model before materialization; each experimental query is then evaluated on the corresponding materialized and SWIP-transformed ProbLog program.

\subsection{MPI3D visual counterfactual program}
\label{app:mpi3d-code}

The MPI3D experiment uses neural predicates for visual factors and deterministic rules for physical predicates.  For each image, the neural predicates return distributions over shape, size, and color.  Counterfactual interventions are then applied to materialized shape or size atoms.

\begin{lstlisting}[style=problogstyle,caption={DeepProbLog template for the MPI3D visual counterfactual experiment.},label={lst:mpi3d-problog}]
% Neural visual predicates.
% For a fixed image I, these are evaluated once and materialized.
nn(shape_net, I, Shape, [cone,cube,cylinder,hexagonal,pyramid,sphere]) :: shape(I, Shape).
nn(size_net,  I, Size,  [small,large]) :: size(I, Size).
nn(color_net, I, Color, [red,green,blue,white,brown,olive]) :: color(I, Color).

% Shape-derived physical properties.
rollable(I) :- shape(I, sphere).
rollable(I) :- shape(I, cylinder).

flat_base(I) :- shape(I, cube).
flat_base(I) :- shape(I, cylinder).
flat_base(I) :- shape(I, hexagonal).

pointed(I) :- shape(I, cone).
pointed(I) :- shape(I, pyramid).

% Size-derived physical properties.
heavy(I) :- size(I, large).
light(I) :- size(I, small).

% Symbolic task predicates.
can_roll(I) :- rollable(I).

can_stack(I) :- flat_base(I).
can_stack(I) :- light(I), \+ pointed(I).

stable(I) :- flat_base(I), heavy(I).
stable(I) :- flat_base(I), light(I).

risky_on_shelf(I) :- rollable(I).
risky_on_shelf(I) :- pointed(I), heavy(I).

% Example queries used after materialization and SWIP transformation.
query(can_roll(I)).
query(can_stack(I)).
query(stable(I)).
query(risky_on_shelf(I)).
\end{lstlisting}

For example, the intervention $do(\mathrm{shape}(I,\mathrm{sphere}))$ removes the materialized mechanism defining the shape of image $I$, redirects downstream uses of shape to the fixed intervention atom, and then evaluates one of the four queries above.  The same materialized program is also used to construct the DeepTwin baseline, so that disagreement would indicate an error in the counterfactual transformation rather than in the neural visual classifiers.

\subsection{SUMO HOV traffic program}
\label{app:sumo-code}

The SUMO HOV experiment uses a neural traffic-state predicate learned from simulated sensor features.  The symbolic program maps traffic state and lane policy to coarse throughput regimes.  The randomized-policy AIPW analysis is not itself a ProbLog program: it is a population estimator applied around the DeepSWIP plug-in nuisance model.  The program below is the symbolic nuisance model whose materialized probabilities are perturbed by sensor noise.

\begin{lstlisting}[style=problogstyle,caption={DeepProbLog template for the SUMO HOV traffic experiment.},label={lst:sumo-problog}]
% Neural traffic-state predicate.
% Sensor features Z are produced by SUMO loop-detector counts and speeds.
nn(traffic_state_net, Z, State, [freeflow,congested]) :: traffic_state(Episode, State).

% Policy atoms.  In the randomized-policy experiment, treatment is assigned
% outside the program and then materialized as one of these facts.
hov_restricted(Episode) :- policy(Episode, restricted).
hov_open(Episode)       :- policy(Episode, open).

% State indicators.
freeflow(Episode)  :- traffic_state(Episode, freeflow).
congested(Episode) :- traffic_state(Episode, congested).

% Coarse symbolic throughput regimes.
high_throughput(Episode) :- freeflow(Episode), hov_open(Episode).
high_throughput(Episode) :- freeflow(Episode), hov_restricted(Episode).

medium_throughput(Episode) :- congested(Episode), hov_open(Episode).
low_throughput(Episode)    :- congested(Episode), hov_restricted(Episode).

% Binary summaries used by the plug-in DeepSWIP nuisance model.
good_outcome(Episode) :- high_throughput(Episode).
good_outcome(Episode) :- medium_throughput(Episode).

% Example evidence and query forms.
% The experiments instantiate these per episode after materialization.
query(good_outcome(Episode)).
query(high_throughput(Episode)).
query(medium_throughput(Episode)).
query(low_throughput(Episode)).
\end{lstlisting}

For the HOV counterfactual query, the intervention $do(\mathrm{policy} = \mathrm{open})$ replaces the policy mechanism with the fixed open-lane policy. Conditioning evidence such as $\mathrm{congested}(i)$ is evaluated in the factual world, while the query is evaluated in the SWIP-transformed program. The AIPW/DML correction reported in the experiment is then applied only to the constructed randomized-policy population estimands, not to arbitrary individual counterfactual queries.

\subsection{Materialized form used by DeepSWIP}
\label{app:materialized-code}

After neural evaluation, each neural predicate is replaced by an ordinary probabilistic choice.  For example, an MPI3D image with shape probabilities $(0.01,0.02,0.05,0.03,0.04,0.85)$ is represented as:

\begin{lstlisting}[style=problogstyle,caption={Example materialized MPI3D neural predicate.},label={lst:materialized-mpi3d}]
0.01::shape(i, cone);
0.02::shape(i, cube);
0.05::shape(i, cylinder);
0.03::shape(i, hexagonal);
0.04::shape(i, pyramid);
0.85::shape(i, sphere).
\end{lstlisting}

Similarly, a SUMO episode with predicted traffic-state probabilities $(0.84,0.16)$ is represented as:

\begin{lstlisting}[style=problogstyle,caption={Example materialized SUMO neural predicate.},label={lst:materialized-sumo}]
0.84::traffic_state(e, freeflow);
0.16::traffic_state(e, congested).
\end{lstlisting}

DeepSWIP is applied only after this materialization step.  This is why the counterfactual solver operates on ordinary ProbLog programs even though the original experimental models contain neural predicates.

The AIPW estimator uses known propensity $g=0.5$ and fold-specific observed-only nuisance estimates:
\begin{equation}
  \widehat\psi_1^{\mathrm{AIPW}}
  =\frac1n\sum_{i=1}^n\left[\hat\mu_1^{(-i)}(W_i)+\frac{T_i}{g}\{Y_i-\hat\mu_1^{(-i)}(W_i)\}\right],
  \label{eq:aipw-psi1}
\end{equation}
with the analogous expression for $\psi_0$ and $\hat\tau=\hat\psi_1-\hat\psi_0$.
\section{Additional Experimental Details}
\label{app:details}

\begin{table}[h]
\centering
\caption{MPI3D visual counterfactuals. SWIP and Twin agree exactly over 12,000 materialized visual-symbolic counterfactual queries. RMSE is against factor-ground-truth potential outcomes.}
\label{tab:mpi3d-results}
\begin{tabular}{lrrrr}
\toprule
Query & SWIP=Twin & Bias & RMSE & Pruning \\
\midrule
\texttt{can\_roll} & 100\% & +0.0000 & 0.0000 & $18\to5$ \\
\texttt{can\_stack} & 100\% & +0.0013 & 0.0580 & $18\to8$ \\
\texttt{risky\_on\_shelf} & 100\% & -0.0026 & 0.0821 & $18\to11$ \\
\texttt{stable} & 100\% & +0.0000 & 0.0000 & $18\to7$ \\
\bottomrule
\end{tabular}
\end{table}

\begin{table}[h]
\centering
\caption{\textbf{SUMO HOV corrected DML/AIPW stress test.}}
\label{tab:hov-results}
\begin{tabular}{rrrrrrrr}
\toprule
$\sigma$ & Cal(ff) & Cal(cg) & Naive $\psi_1$ & CV $\psi_1$ & AIPW $\psi_1$ & Naive ATE & AIPW ATE \\
\midrule
0.0 & 0.920 & 0.985 & -4.75 & -4.67 & +0.14 & -1.24 & +0.15 \\
1.0 & 0.890 & 0.900 & -5.66 & -5.36 & +0.08 & -1.50 & +0.06 \\
2.0 & 0.845 & 0.706 & -6.32 & -7.04 & +0.13 & -1.68 & +0.14 \\
3.0 & 0.767 & 0.556 & -9.55 & -8.92 & +0.03 & -2.57 & -0.03 \\
\bottomrule
\end{tabular}
\end{table}

\end{document}